# Credal Networks under Maximum Entropy

**Thomas Lukasiewicz**
Institut und Ludwig Wittgenstein Labor für Informationssysteme, TU Wien
Favoritenstraße 9-11, A-1040 Vienna, Austria
lukasiewicz@kr.tuwien.ac.at

## Abstract

We apply the principle of maximum entropy to select a unique joint probability distribution from the set of all joint probability distributions specified by a credal network. In detail, we start by showing that the unique joint distribution of a Bayesian tree coincides with the maximum entropy model of its conditional distributions. This result, however, does not hold anymore for general Bayesian networks. We thus present a new kind of maximum entropy models, which are computed sequentially. We then show that for all general Bayesian networks, the sequential maximum entropy model coincides with the unique joint distribution. Moreover, we apply the new principle of sequential maximum entropy to interval Bayesian networks and more generally to credal networks. We especially show that this application is equivalent to a number of small local entropy maximizations.

## 1 INTRODUCTION

In classical Bayesian networks [31], a single joint probability distribution is specified by a number of conditional independencies encoded in a directed acyclic graph and a number of single conditional probability distributions. More precisely, we assume in particular that all conditional probability distributions are precisely given.

There are, however, several reasons for which we should give up such an assumption. In realistic environments, we often deal with interval rather than point probabilities. Moreover, an agent may be unable to specify precise probabilities due to limited resources of information or time; or an agent just wants a rough analysis of a decision situation. Finally, it may be the case that the decision situation itself cannot be modeled by precise probabilities.

The need for giving up precise probabilities in Bayesian networks is already reported by a number of publications in this direction. Intervals in the Bayesian network framework are especially discussed by Breese and Fertig [2], Tessem [37], Thöne et al. [38], and Zaffalon [12]. Bayesian networks with local convex sets of conditional distributions, called *credal networks* (or also *Quasi-Bayesian networks*), are especially analyzed by Cozman [7, 8].

One philosophy to handle credal networks is to insist on working with *sets of probability distributions*. In this case, an agent follows a very cautious path by considering every distribution as relevant. But this also means that there might be ambiguous situations in which the agent simply does not know what to do. Moreover, following this philosophy also means developing completely new propagation and inference techniques for credal networks (see especially the work by Cano et al. [3, 4] and Cozman [7] for exact and approximation techniques). To our knowledge, current exact propagation techniques for credal networks do not go much beyond interval Bayesian polytrees over binary random variables [38, 12].

Another philosophy is to select a representative probability distribution from the set of all specified distributions. In this case, an agent follows a credulous path, as the selected distribution might be the wrong one. Nevertheless, the agent does not have to deal with ambiguities resulting from multiple distributions. Moreover, after selecting the unique representative distribution, we can simply apply classical Bayesian network techniques.

In this paper, we adhere to the second philosophy. In detail, we propose to use the principle of *sequential maximum entropy* to select a representative distribution from the set of all distributions specified by a credal network.

We now describe the main ideas behind this proposal.

A Bayesian network $(D, KB)$ is given by a directed acyclic graph $D$ and a set of conditional distributions $KB$. It defines a unique joint probability distribution $Pr_D[KB]$ by the conditional distributions $KB$ and certain conditional independencies encoded in $D$.

Interestingly, it turns out that for Bayesian trees $(D, KB)$, the unique joint distribution $Pr_D[KB]$ coincides with the maximum entropy model of $KB$ (which is the unique joint distribution that is compatible with all conditional distributions in $KB$ and that has the greatest entropy among all such joint distributions). That is, the maximum entropy model of $KB$ naturally respects all the conditional independencies encoded in $D$.

This suggests that maximum entropy may be used to select a unique joint distribution from the set of all joint distributions $Pr_D[KB]$ specified by a credal network $(D, KB)$.

Indeed, we will see that also for credal trees $(D, KB)$, the maximum entropy model of $KB$ naturally respects all conditional independencies encoded in $D$.

But, for general Bayesian and credal networks $(D, KB)$, the maximum entropy model of $KB$ does not necessarily respect all the conditional independencies encoded in $D$. This result is well-known in the literature as the *causality problem* of minimum cross-entropy updating [31, 16].

Moreover, also for credal trees $(D, KB)$, selecting the maximum entropy model of $KB$ as a representative joint distribution may contradict intuition. In detail, we may have *causality problems on a meta-level*.

Generalizing an idea that goes back to Hunter [16], we then introduce the *principle of sequential maximum entropy* as a selection principle that solves our two causality problems. More precisely, for all Bayesian networks $(D, KB)$, the sequential maximum entropy model of $KB$ coincides with the unique joint distribution $Pr_D[KB]$. Moreover, for all credal networks $(D, KB)$, the sequential maximum entropy model of $KB$ respects all the conditional independencies encoded in $D$. Finally, it also turns out that the sequential maximum entropy model can be computed by a number of small local entropy maximizations.

The main contributions of this paper are as follows:

- We present an I-map for the maximum entropy model of a set of convex conditionals. Such an I-map is a very useful characterization of conditional independencies, which is important in its own right.
- We show that for all Bayesian trees $(D, KB)$, the maximum entropy model of $KB$ coincides with the unique joint distribution defined by $KB$ and the conditional independencies encoded in $D$.
- We show that for all credal trees $(D, KB)$, the maximum entropy model of $KB$ respects all conditional independencies encoded in $D$.
- We introduce the principle of sequential maximum entropy, which can be applied to all credal networks.
- We show that for all Bayesian networks $(D, KB)$, the sequential maximum entropy model of $KB$ coincides with the unique joint distribution of $(D, KB)$.

- We show that for all credal networks $(D, KB)$, the sequential maximum entropy model of $KB$ respects all conditional independencies encoded in $D$.
- We show that the sequential maximum entropy model of a credal network can be computed by a number of small local entropy maximizations.

The rest of this paper is organized as follows. Section 2 introduces the technical background. In Section 3, we concentrate on Bayesian and credal networks under global maximum entropy. Sections 4 and 5 focus on the principle of sequential maximum entropy. In Section 6, we finally give a summary, and an outlook on future research.

Note that all proofs are given in full detail in [24].

## 2 TECHNICAL PRELIMINARIES

In this section, we describe the technical background.

### 2.1 CREDAL NETWORKS

We now give a brief introduction to Bayesian networks, interval Bayesian networks, and credal networks (see especially [31], [37, 38, 12], and [7, 8], respectively).

A *Bayesian network* is defined by a directed acyclic graph $D$ over discrete random variables $X_1, X_2, \ldots, X_n$ as nodes and by a conditional probability distribution $Pr(X_i | \mathbf{pa}(X_i))$ for each variable $X_i$ and each instantiation $\mathbf{pa}(X_i)$ of its parents $pa(X_i)$. It specifies a unique joint probability distribution $Pr$ over $X_1, X_2, \ldots, X_n$ by:

$$Pr(X_1, X_2, \ldots, X_n) \;=\; \prod_{i=1}^{n} Pr(X_i | pa(X_i))\,.$$

That is, the joint distribution $Pr$ is uniquely determined by the conditional distributions $Pr(X_i | \mathbf{pa}(X_i))$ and certain conditional independencies encoded in $D$.

More generally, a *credal network* is defined by a directed acyclic graph $D$ over $X_1, X_2, \ldots, X_n$ and by a nonempty convex set $\mathcal{S}(X_i | \mathbf{pa}(X_i))$ of conditional probability distributions $Pr(X_i | \mathbf{pa}(X_i))$ for each variable $X_i$ and each instantiation $\mathbf{pa}(X_i)$ of its parents $pa(X_i)$.

We associate a credal network with the set of all joint distributions that are admissible with the convex sets of conditional distributions and the conditional independencies encoded in $D$ [12]. See in particular [8] for other possible semantics of credal networks (especially those that involve new notions of irrelevance and independency).

*Interval Bayesian networks* are a special kind of credal networks in which each $\mathcal{S}(X_i | \mathbf{pa}(X_i))$ can be expressed by a set of interval constraints $\mathcal{I}(X_i | \mathbf{pa}(X_i))$ of the form

$$\{l_j \leq Pr(X_i = x_{i,j} | \mathbf{pa}(X_i)) \leq u_j \mid j \in [1 : d_i]\}\,,$$

where $\{x_{i,1}, \ldots, x_{i,d_i}\}$ with $d_i \geq 2$ denotes the domain of the variable $X_i$, $0 \leq l_j \leq u_j \leq 1$ for all $j \in \{1, \ldots, d_i\}$, and $l_1 + \cdots + l_{d_i} \leq 1 \leq u_1 + \cdots + u_{d_i}$.

A Bayesian (resp., interval Bayesian, credal) network is a *Bayesian (resp., interval Bayesian, credal) tree* iff the associated undirected graph $D$ is a directed tree (that is, a directed acyclic graph in which every node has exactly one incoming arrow, except for the *root* that does not have any).

## 2.2 CONDITIONALS

We will use the language of conditionals (see especially [13, 21, 25, 26]) to represent Bayesian networks, interval Bayesian networks, and credal networks.

Let $U = \{X_1, \ldots, X_n\}$ with $n \geq 1$ be a set of discrete random variables, where each variable $X_i \in U$ has a finite and nonempty domain $D_{X_i} = \{x_{i,1}, \ldots, x_{i,d_i}\}$. The set of *basic events* $\mathcal{B}_U$ contains all expressions of the form $X_i = x_i$ with $X_i \in U$ and $x_i \in D_{X_i}$. The set of *conjunctive events* $\mathcal{C}_U$ contains the *true event* $\top$ and all members in the closure of $\mathcal{B}_U$ under the Boolean operation $\wedge$. We abbreviate the conjunctive event $c \wedge d$ by $c, d$. An *instantiation* of a set of variables $\{X_{i_1}, \ldots, X_{i_k}\} \subseteq U$ with $k \geq 1$ is a conjunctive event of the form $X_{i_1} = x_{i_1}, \ldots, X_{i_k} = x_{i_k}$ (the unique instantiation of the empty set of variables is $\top$).

A *point conditional* is an expression of the form $(d\,|\,c)[r]$, where $c$ and $d$ are conjunctive events (called *premise* and *conclusion*, respectively) and $r$ is a real number from $[0, 1]$. An *interval conditional* is an expression $(d\,|\,c)[l, u]$ with conjunctive events $c$ and $d$ and real numbers $l, u \in [0, 1]$ such that $l \leq u$. A *convex conditional* is an expression $(X_i\,|\,c)[K]$ with a variable $X_i \in U$, and a finitely generated nonempty convex set $K \subseteq [0, 1]^{d_i}$ such that $r_1 + \cdots + r_{d_i} = 1$ for all $(r_1, \ldots, r_{d_i}) \in K$ (note that $K$ is *finitely generated* iff it is the convex closure of a finite number of vectors $\vec{k}_1, \ldots, \vec{k}_l \in [0, 1]^{d_i}$ with $l \geq 0$). A *conditional* is a point, interval, or convex conditional.

To define probabilistic interpretations of conjunctive events and conditionals, we introduce atomic events and the binary relation '$\Rightarrow$' between atomic and conjunctive events. The set of *atomic events* $\Omega_U$ contains all conjunctive events of the form $X_1 = x_1, X_2 = x_2, \ldots, X_n = x_n$. The atomic event $\omega$ *implies* the conjunctive event $c$, denoted $\omega \Rightarrow c$, iff $\omega \wedge \neg c$ is a propositional contradiction (that is, each basic event in $c$ is also contained in $\omega$).

A *probabilistic interpretation* $Pr$ is a mapping from $\Omega_U$ to $[0, 1]$ such that all $Pr(\omega)$ with $\omega \in \Omega_U$ sum up to 1. $Pr$ is extended in a well-defined way to conjunctive events $c$ by defining $Pr(c)$ as sum of all $Pr(\omega)$ with $\omega \in \Omega_U$ and $\omega \Rightarrow c$. For conjunctive events $c$ and $d$ with $Pr(c) > 0$, we write $Pr(d\,|\,c)$ to abbreviate $Pr(c, d) / Pr(c)$. A probabilistic interpretation $Pr$ is extended to point conditionals by $Pr \models (d\,|\,c)[r]$ iff $Pr(c) = 0$ or $Pr(d\,|\,c) = r$. Moreover,

$Pr$ is extended to interval conditionals by $Pr \models (d\,|\,c)[l, u]$ iff $Pr(c) = 0$ or $Pr(d\,|\,c) \in [l, u]$. Finally, $Pr$ is extended to convex conditionals as follows:

$$Pr \models (X_i\,|\,c)[K] \text{ iff}$$
$$(Pr(X_i = x_{i,1}\,|\,c), \ldots, Pr(X_i = x_{i,d_i}\,|\,c)) \in K$$

The notions of models and satisfiability are defined as usual: An interpretation $Pr$ is a *model* of a conditional $F$ iff $Pr \models F$. $Pr$ is a *model* of a set of conditionals $KB$, denoted $Pr \models KB$, iff $Pr$ is a model of all $F \in KB$. A set of conditionals $KB$ is *satisfiable* iff a model of $KB$ exists.

## 2.3 CREDAL NETWORKS AS CONDITIONALS

We now express Bayesian networks, interval Bayesian networks, and credal networks by sets of conditionals.

A *Bayesian network* is a pair $(D, KB)$, where $D$ is a directed acyclic graph over $U$ as nodes and $KB$ is a set of point conditionals such that:

- For each variable $X_i \in U$ and each instantiation $\mathbf{pa}(X_i)$ of its parents $\mathrm{pa}(X_i)$, there exists a set of conditionals $KB_{X_i\,|\,\mathbf{pa}(X_i)}$ of the form

$$\{(X_i = x_{i,j}\,|\,\mathbf{pa}(X_i))[r_j]\,|\,j \in [1 : d_i]\}$$

with $r_1 + \cdots + r_{d_i} = 1$.

- $KB$ is the union of all $KB_{X_i\,|\,\mathbf{pa}(X_i)}$.

Given a Bayesian network $(D, KB)$, we use $Pr_D[KB]$ to denote its unique joint probability distribution.

Similarly, an *interval Bayesian network* is a pair $(D, KB)$, where $D$ is a directed acyclic graph over $U$ as nodes and $KB$ is a set of interval conditionals such that:

- For each variable $X_i \in U$ and each instantiation $\mathbf{pa}(X_i)$ of its parents $\mathrm{pa}(X_i)$, there exists a set of interval conditionals $KB_{X_i\,|\,\mathbf{pa}(X_i)}$ of the form

$$\{(X_i = x_{i,j}\,|\,\mathbf{pa}(X_i))[l_j, u_j]\,|\,j \in [1 : d_i]\}$$

with $l_1 + \cdots + l_{d_i} \leq 1 \leq u_1 + \cdots + u_{d_i}$.

- $KB$ is the union of all $KB_{X_i\,|\,\mathbf{pa}(X_i)}$.

Finally, a *credal network* is a pair $(D, KB)$, where $D$ is a directed acyclic graph over $U$ as nodes and $KB$ is a set of convex conditionals such that:

- For each variable $X_i \in U$ and each instantiation $\mathbf{pa}(X_i)$ of its parents $\mathrm{pa}(X_i)$, there exists a set of convex conditionals $KB_{X_i\,|\,\mathbf{pa}(X_i)}$ of the form

$$\{(X_i\,|\,\mathbf{pa}(X_i))[K]\}.$$

- $KB$ is the union of all $KB_{X_i|\mathbf{pa}(X_i)}$.

Given an interval Bayesian network $(D, KB)$ or a credal network $(D, KB)$, we use $Pr_D[KB]$ to denote the set of all associated joint probability distributions.

## 2.4 MAXIMUM ENTROPY MODELS

We now define maximum entropy models of sets of conditionals. For the principle of maximum entropy see especially the work by Shannon and Weaver [35] and Jaynes [17]. Its application to probabilistic reasoning in the artificial intelligence context are in particular discussed by Cheeseman [6], Paris and Vencovska [29, 30], and recently also by Grove et al. [14].

The *maximum entropy model (ME-model)* of a satisfiable set of conditionals $KB$, denoted $Pr_{ME}[KB]$, is the unique probabilistic interpretation $Pr$ that is a model of $KB$ and that has the greatest entropy $H(Pr)$ among all the models of $KB$, where $H(Pr)$ is defined as follows:

$$H(Pr) \; = \; - \sum_{\omega \in \Omega_U} Pr(\omega) \cdot \log Pr(\omega) \, .$$

## 2.5 I-MAPS OF MAXIMUM ENTROPY MODELS

As a first contribution of this paper, we now describe how to construct an I-map for the ME-model of a satisfiable set of conditionals $KB$ (that is, a characterization of conditional independencies holding in $Pr_{ME}[KB]$). Similar constructions of undirected graphs from formulas are given in [22] and [33], which produce I-maps for probability distributions under conditioning, and *dependency graphs* for ME-models of sets of point conditionals, respectively.

Let us briefly recall that for pairwise disjoint sets of variables $X, Y, Z \subseteq U$, we say $X$ and $Y$ are *conditionally independent* given $Z$ in $Pr$ iff for all instantiations $x$, $y$, and $z$ of $X$, $Y$, and $Z$, respectively:

$$Pr(x|y, z) \; = \; Pr(x|z) \text{ whenever } Pr(y, z) > 0 \, .$$

An undirected graph $(U, E)$ is an I-map of a probabilistic interpretation $Pr$ iff for all pairwise disjoint $X, Y, Z \subseteq U$: if $X$ and $Y$ are separated by $Z$ in $(U, E)$, then $X$ and $Y$ are conditionally independent given $Z$ in $Pr$.

The main idea in building an I-map of $Pr_{ME}[KB]$ is to add an edge between two different variables iff there is a conditional in $KB$ that contains them both.

More formally, the undirected graph $G_{KB} = (U, E)$ contains the undirected edge $\{X, Y\}$ iff $X$ and $Y$ are two different variables that both occur in the same conditional $(d|c)[r] \in KB$, $(d|c)[l, u] \in KB$, or $(D|c)[K] \in KB$ That is, both $X$ and $Y$ occur in $d$, both occur in $c$, or one of them occurs in $d$ (resp., $D$) and the other one in $c$.

The following result shows that the constructed undirected graph $G_{KB}$ is indeed an I-map of the ME-model of $KB$.

**Theorem 2.1** *Let $KB$ be a satisfiable set of conditionals. Then, $G_{KB}$ is an I-map of $Pr_{ME}[KB]$.*

# 3   GLOBAL MAXIMUM ENTROPY

In this section, we concentrate on Bayesian and credal networks under global maximum entropy.

## 3.1   BAYESIAN NETWORKS

We first analyze the relationship between Bayesian networks and global maximum entropy. In detail, given a Bayesian network $(D, KB)$, we study the relationship between the conditional independencies encoded in $D$ and the conditional independencies in the ME-model of $KB$.

Interestingly, as far as Bayesian trees $(D, KB)$ are concerned, all the conditional independencies encoded in $D$ also hold in the ME-model of $KB$ (that is, the conditional independencies in $D$ are naturally entrenched in $KB$).

**Theorem 3.1** *For all Bayesian trees $(D, KB)$, it holds $Pr_{ME}[KB] = Pr_D[KB]$.*

This remarkable result, however, does not carry over to general Bayesian networks $(D, KB)$. In this more general case, the ME-model of $KB$ does not necessarily respect all the conditional independencies encoded in $D$. This shows the following theorem, which is well-known as the *causality problem* of minimum cross-entropy updating [31, 16].

**Theorem 3.2 (essentially [31, 16])** *There exist Bayesian networks $(D, KB)$ such that $Pr_{ME}[KB] \neq Pr_D[KB]$.*

## 3.2   CREDAL NETWORKS

We now generalize the results of the previous section to credal networks $(D, KB)$. Clearly, in the general case, by Theorem 3.2, the ME-model of $KB$ does not necessarily respect all the conditional independencies encoded in $D$.

In the special case of credal trees $(D, KB)$, however, the independencies of $D$ are also respected in the ME-model of $KB$. That is, Theorem 3.1 carries over to credal trees.

**Theorem 3.3** *For all credal trees $(D, KB)$, it holds $Pr_{ME}[KB] \in Pr_D[KB]$.*

This result suggests that maximum entropy can be used to select a unique joint distribution from the set of all joint distributions specified by a credal tree. The next example, however, shows that this selection may be counter-intuitive.

**Example 3.1** Let the directed acyclic graph $D$ over the discrete random variables $A$, $B$, and $C$ with the domains $\{a, \bar{a}\}$, $\{b, \bar{b}\}$, and $\{c, \bar{c}\}$, respectively, be given by Fig. 1.

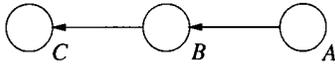

Figure 1: Directed Tree $D$

Assume that $A = a$, $B = b$, and $C = c$ represent the events "burglary", "alarm sound", and "phone call", respectively, of the classical burglary example [31] in which Mr. Holmes has to deal with the following scenario. A burglary at his house would probably start the alarm system, which itself would probably cause his daughter to give him a phone call. Moreover, suppose that Mr. Holmes does not know anything else than the probability $u \in [0, 1]$ that his daughter will give him a phone call when she hears the alarm sound. That is, assume $KB_u = \{(C = c \mid B = b)[u]\}$.

How does the ME-model of $KB_u$ now look like? For instance, which is the conditional probability of alarm sound given a burglary, that is, $Pr_{ME}[KB_u](B = b \mid A = a)$?

Interestingly, this conditional probability strongly depends on $u$ as shown in Fig. 2 (it is given by the function $f(u) = 1 / (1 + 2 u^u (1 - u)^{(1-u)})$). That is, under global maximum entropy, the selected probability that the alarm starts when there is a burglary depends on the probability that the daughter calls when she hears the alarm sound.

This seems highly counter-intuitive.

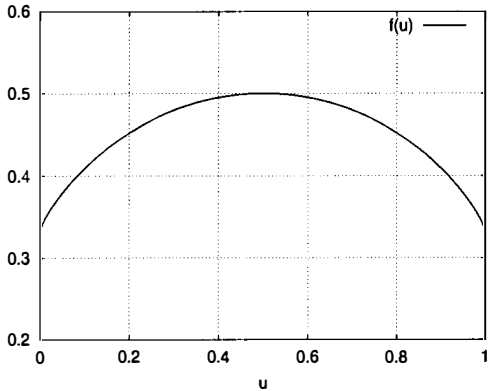

Figure 2: $f(u) = Pr_{ME}[KB_u](B = b \mid A = a)$

## 4    SEQUENTIAL MAXIMUM ENTROPY

In this section, we present the principle of sequential maximum entropy for selecting unique joint distributions among all joint distributions specified by a credal network.

### 4.1    MAIN IDEAS

As shown by Theorem 3.2 and Example 3.1, there are two main problems coming along with applying the principle of maximum entropy to Bayesian and credal networks.

The first problem is the well-known improper handling of causal information. That is, maximum entropy produces undesired dependencies between the variables in the premises of conditionals (see Theorem 3.2).

The second problem is that the principle of maximum entropy also produces undesired dependencies between conditionals on a meta-level (see Example 3.1).

The first problem has already been addressed in the literature. In detail, Hunter [16] proposed to use probabilistic counterfactuals as a more adequate representation of causal conditional probability statements. He shows that maximum entropy applied to such probabilistic counterfactuals handles causal information in a correct way.

More precisely, Hunter considers the restricted case of a Bayesian network over the directed acyclic graph

$$D = (\{C, A_1, \ldots, A_k\}, \{C \leftarrow A_i \mid i \in \{1, \ldots, k\})$$

with binary random variables $C, A_1, \ldots, A_k$ and $k \geq 2$. He represents each conditional distribution $Pr(C \mid \mathbf{pa}(C))$ by a probabilistic counterfactual. Such probabilistic counterfactuals are interpreted by probability distributions over the set of all linear orders of atomic events. Hunter then shows that minimum cross-entropy updating of a prior distribution w.r.t. the given set of probabilistic counterfactuals does not introduce any new dependencies between $A_1, \ldots, A_k$. Finally, he remarks that in certain cases, the same effect can be obtained by keeping $Pr(A_1, \ldots, A_k)$ fixed to its values in the prior distribution and then performing minimum cross-entropy updating w.r.t. all $Pr(C \mid \mathbf{pa}(C))$.

This remark will be the first important building block of our principle of *sequential maximum entropy*, which can be applied to all credal networks $(D, KB)$. In detail, for each variable $X_i$ we will perform a maximum entropy computation with respect to all $Pr(X_i \mid \mathbf{pa}(X_i))$, while keeping previously computed probability values fixed.

That is, we now just have to determine the *linear order* in which these single computations must be done. We will see that we can use any ordering $(X_1, X_2, \ldots, X_n)$ of the variables in $U$ that respects the structure of $D$. This *sequential* computation will then also solve our second problem with global maximum entropy reported by Example 3.1.

We remark, however, that the sequential maximum entropy model of $KB$ clearly depends on the set of variables in $U$, their domains of values, and the structure of $D$.

Finally, one intuition behind sequential maximum entropy can informally be described as follows. Assume that each arrow $X \rightarrow Y$ in $D$ represents a temporal relationship in the sense that any event related to $X$ happens temporally before any event related to $Y$. Hence, at any variable $Z$, we cannot change anymore the probabilities of instantiations of variables "in the past", and we are not influenced by any probabilities related to variables "in the future".

## 4.2 BAYESIAN NETWORKS

We now introduce the principle of sequential maximum entropy for Bayesian networks. We will show that for all Bayesian networks $(D, KB)$, the sequential maximum entropy model of $KB$ coincides with the unique joint distribution of $(D, KB)$ (that is, the sequential maximum entropy model respects all the conditional independencies in $D$).

The main idea is to apply the principle of maximum entropy in an sequential way. In detail, we take an ordering $(X_1, X_2, \ldots, X_n)$ of the variables in $U$ that is consistent with $D$ (that is, if $X_i \rightarrow X_j$ is an arrow in $D$, then $i < j$). In the sequel, we assume that the indices of the variables in $U = \{X_1, X_2, \ldots, X_n\}$ already respect this ordering.

We now compute ME-models with respect to each set of variables $U_i = \{X_1, X_2, \ldots, X_i\}$, where $i$ is increasing from 1 to $n$. In each iteration step $i$, we involve exactly those conditionals in the maximum entropy computation that have $X_i$ in their conclusion, and we also involve new conditionals that keep fixed all what we computed so far in previous iteration steps. Note that these new conditionals are crucial, since otherwise each maximum entropy computation would destroy the results of previous computations by assuming dependencies where we do not want any.

More formally, the *sequential maximum entropy model (sequential ME-model)* of a Bayesian network $(D, KB)$, denoted $Pr_{ME}^{seq}[KB]$, is defined as $Pr_{ME}[KB_n]$, where $KB_1$ and $KB_i$ for $1 < i \leq n$ are defined as follows:

$$KB_1 = KB_{X_1 | \top}$$
$$KB_i = \bigcup_{\mathbf{pa}(X_i)} KB_{X_i | \mathbf{pa}(X_i)} \cup$$
$$\{(\omega \mid \top)[Pr_{ME}[KB_{i-1}](\omega)] \mid \omega \in \Omega_{U_{i-1}}\}.$$

The following result shows that the sequential ME-model respects all the conditional independencies encoded in $D$ (note that this also shows that the sequential ME-model does not depend on the selected ordering of the variables).

**Theorem 4.1** *Let $(D, KB)$ be a Bayesian network. Then, it holds $Pr_{ME}^{seq}[KB] = Pr_D[KB]$.*

## 4.3 CREDAL NETWORKS

We now apply the principle of sequential maximum entropy to interval Bayesian networks and credal networks.

The *sequential ME-model* of an interval Bayesian (resp., credal) network is defined exactly like the sequential ME-model of a classical Bayesian network (see Section 4.2).

### 4.3.1 Interval Bayesian Networks

We first focus on interval Bayesian networks. The next result shows that their sequential ME-model can be computed by local entropy maximizations (one for each variable $X_i \in U$ and each instantiation of its parents $\mathbf{pa}(X_i)$).

**Theorem 4.2** *Let $(D, KB)$ be an interval Bayesian network. Let the Bayesian network $(D, KB^\star)$ be built from $(D, KB)$ by replacing each set of interval conditionals*

$$KB_{X_i | \mathbf{pa}(X_i)} = \{(X_i = x_{i,j} \mid \mathbf{pa}(X_i))[l_j, u_j] \mid j \in [1 : d_i]\}$$

*by the new set of conditionals*

$$KB_{X_i | \mathbf{pa}(X_i)}^\star = \{(X_i = x_{i,j} \mid \mathbf{pa}(X_i))[r_j^\star] \mid j \in [1 : d_i]\},$$

*where $(r_1^\star, \ldots, r_{d_i}^\star)$ is the optimal solution of the following optimization problem (over $r_1, \ldots, r_{d_i} \geq 0$):*

$$\max \sum_{j=1}^{d_i} -r_j \log r_j \quad \text{subject to } \mathcal{LC}_I \qquad (1)$$

*with $\mathcal{LC}_I$ being the least set of linear constraints containing $r_1 + \cdots + r_{d_i} = 1$ and $l_j \leq r_j \leq u_j$ for all $j \in [1 : d_i]$.*

*Then, it holds $Pr_D[KB^\star] = Pr_{ME}^{seq}[KB]$.*

An immediate corollary is that the sequential ME-model respects all the conditional independencies encoded in $D$.

**Corollary 4.3** *Let $(D, KB)$ be an interval Bayesian network. Then, it holds $Pr_{ME}^{seq}[KB] \in Pr_D[KB]$.*

Moreover, the sequential ME-model does not depend on the selected ordering of the variables.

**Corollary 4.4** *Let $(D, KB)$ be an interval Bayesian network. Then, the same sequential ME-model of $KB$ is obtained for every ordering $(X_1, \ldots, X_n)$ of the variables in $U$ that is consistent with $D$.*

Summarizing, the sequential ME-model of an interval Bayesian network can be computed as follows. For each variable $X_i \in U$ and each instantiation of its parents $\mathbf{pa}(X_i)$, we perform one local entropy maximization over $d_i$ variables subject to $2d_i + 1$ linear constraints.

Such local entropy maximizations can easily be done by existing maximum entropy software. For example, the system SPIRIT [33] works with point conditionals, and can thus be used to handle *incomplete Bayesian networks* (that is, interval Bayesian networks that contain only conditionals of the form $(d|c)[r]$ or $(d|c)[0, 1]$). Moreover, PIT [34, 11] also works with interval conditionals, and can thus be used for interval Bayesian networks in their full generality.

### 4.3.2 General Credal Networks

We now generalize the results of Section 4.3.1 to credal networks. The next theorem shows that the sequential ME-model can be computed by local entropy maximizations.

**Theorem 4.5** *Let $(D, KB)$ be a credal network. Let the Bayesian network $(D, KB^\star)$ be built from $(D, KB)$ by replacing each set of convex conditionals*

$$KB_{X_i | \mathbf{pa}(X_i)} = \{(X_i \mid \mathbf{pa}(X_i))[K]\}$$

by the new set of conditionals

$$KB^*_{X_i|\mathbf{pa}(X_i)} = \{(X_i=x_{i,j} \mid \mathbf{pa}(X_i))[r^*_j] \mid j \in [1:d_i]\},$$

where $(r^*_1, \ldots, r^*_{d_i})$ is the optimal solution of the following optimization problem (over $r_1, \ldots, r_{d_i} \geq 0$):

$$\max \sum_{j=1}^{d_i} -r_j \log r_j \text{ subject to } (r_1, \ldots, r_{d_i}) \in K. \quad (2)$$

Then, it holds $Pr_D[KB^*] = Pr^{seq}_{ME}[KB]$.

It immediately follows that the sequential ME-model respects all the conditional independencies encoded in $D$.

**Corollary 4.6** For all credal networks $(D, KB)$, it holds $Pr^{seq}_{ME}[KB] \in Pr_D[KB]$.

Moreover, the sequential ME-model does not depend on the selected ordering of the variables.

**Corollary 4.7** Let $(D, KB)$ be a credal network. Then, the same sequential ME-model of $KB$ is obtained for every ordering $(X_1, X_2, \ldots, X_n)$ of the variables in $U$ that is consistent with $D$.

Hence, if we express convex sets by linear constraints, then the sequential ME-model can be computed as follows. For each convex conditional $(X_i \mid \mathbf{pa}(X_i))[K] \in KB$, we perform one local entropy maximization over $d_i$ variables subject to the linear constraints that represent $K$.

# 5 EXAMPLES

In this section, we give some examples to illustrate the principle of sequential maximum entropy.

**Example 5.1** Consider again the interval Bayesian network $(D, KB_u)$ of Example 3.1. Under sequential maximum entropy, the selected probability that the alarm starts when there is a burglary *does not depend anymore* on the probability that the daughter calls when she hears the alarm sound. More precisely, $Pr^{seq}_{ME}[KB_u](B=b \mid A=a)$ is given by 0.5 and thus independent of $u$.

In the next example, we consider an interval Bayesian network with cycles and non-binary random variables.

**Example 5.2** Let the directed acyclic graph $D$ over the discrete random variables $A$, $B$, $C$, $E$, and $F$ with the domains $\{a_1, a_2\}$, $\{b_1, b_2\}$, $\{c_1, c_2\}$, $\{e_1, e_2\}$, and $\{f_1, f_2, f_3\}$, respectively, be given by Fig. 3.

Some entailed conditionals in an interval Bayesian network $(D, KB)$ and their corresponding conditionals in the Bayesian network $(D, KB^*)$ produced by the principle of sequential maximum entropy are shown in Table 1.

Let us consider some entailed conditional probabilities. For instance, $Pr^{seq}_{ME}[KB](F=f_1 \mid A=a_1) = 0.64$, while the minimum (resp., maximum) of $Pr(F=f_1 \mid A=a_1)$ subject to $Pr \in Pr_D[KB]$ is given by 0.63 (resp., 0.84).

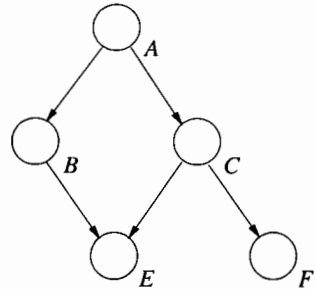

Figure 3: Directed Tree $D$

Table 1: Some Conditionals in $KB$ and $KB^*$

|  | $KB$ | $KB^*$ |
|---|---|---|
| $(A=a_1 \mid \top)$ | $[0.2, 0.7]$ | $[0.5]$ |
| $(A=a_2 \mid \top)$ | $[0.3, 0.8]$ | $[0.5]$ |
| $(C=c_1 \mid A=a_1)$ | $[0.3, 0.4]$ | $[0.4]$ |
| $(C=c_2 \mid A=a_1)$ | $[0.6, 0.7]$ | $[0.6]$ |
| $(C=c_1 \mid A=a_2)$ | $[0.3, 0.5]$ | $[0.5]$ |
| $(C=c_2 \mid A=a_2)$ | $[0.5, 0.7]$ | $[0.5]$ |
| $(F=f_1 \mid C=c_1)$ | $[0.7, 0.9]$ | $[0.7]$ |
| $(F=f_2 \mid C=c_1)$ | $[0.0, 0.3]$ | $[0.15]$ |
| $(F=f_3 \mid C=c_1)$ | $[0.0, 0.3]$ | $[0.15]$ |
| $(F=f_1 \mid C=c_2)$ | $[0.6, 0.8]$ | $[0.6]$ |
| $(F=f_2 \mid C=c_2)$ | $[0.1, 0.3]$ | $[0.2]$ |
| $(F=f_3 \mid C=c_2)$ | $[0.1, 0.3]$ | $[0.2]$ |

# 6 SUMMARY AND OUTLOOK

We showed that the unique joint distribution of a Bayesian tree coincides with the maximum entropy model of its conditional distributions. We then presented a new kind of maximum entropy models, which are computed sequentially. We showed that for all general Bayesian networks, the sequential maximum entropy model coincides with the unique joint distribution. We then applied the new principle of sequential maximum entropy to credal networks. We especially showed that this application is equivalent to a number of small local entropy maximizations.

A very interesting topic of future research is to apply the results of this work to the framework of probabilistic logic programming [32, 15, 28]. Moreover, it would be interesting to use the principle of sequential maximum entropy in order to add causality to probabilistic default reasoning with conditional constraints [27].

## Acknowledgments

I am very grateful to Fabio Gagliardi Cozman, Gabriele Kern-Isberner, and Richard Neapolitan for their useful comments on an earlier version of this paper. Many thanks also to the referees for their useful suggestions.

This work has been partially supported by a DFG grant and the Austrian Science Fund Project N Z29-INF.